\documentclass[10pt,letterpaper]{article}
\usepackage[top=0.85in,left=2.75in,footskip=0.75in]{geometry}

\usepackage{amsmath,amssymb}

\usepackage{changepage}

\usepackage[utf8x]{inputenc}

\usepackage{textcomp,marvosym}

\usepackage{cite}

\usepackage{nameref,hyperref}

\usepackage[right]{lineno}

\usepackage{microtype}
\DisableLigatures[f]{encoding = *, family = * }

\usepackage[table]{xcolor}

\usepackage{array}

\usepackage{enumerate}
\usepackage{enumitem}
\usepackage{multicol}

\newcolumntype{+}{!{\vrule width 2pt}}

\newlength\savedwidth

\raggedright
\usepackage[aboveskip=1pt,labelfont=bf,labelsep=period,justification=raggedright,singlelinecheck=off]{caption}

\makeatletter
\renewcommand{\@biblabel}[1]{\quad#1.}
\makeatother

\usepackage{lastpage,fancyhdr,graphicx}
\usepackage{epstopdf}
\fancyheadoffset[L]{2.25in}
\fancyfootoffset[L]{2.25in}
\newcommand{\Cifar}{\textsc{Cifar10}}
\definecolor{Gray}{gray}{0.85}

\begin{document}
\vspace*{0.2in}

\begin{flushleft}
{\Large
\textbf\newline{Entropic associative memory for real world images} 
}
\newline
\\
Noé Hernández\textsuperscript{1,2*},
Rafael Morales\textsuperscript{3},
Luis A. Pineda\textsuperscript{2}
\\
\bigskip
\textbf{1} Posgrado en Ciencia e Ingeniería de la Computación, UNAM, México
\\
\textbf{2} Universidad Nacional Autónoma de México, IIMAS, 04510, México
\\
\textbf{3} Universidad de Guadalajara, SUV, Guadalajara, 44130, México
\\
\bigskip

%
%





* nohernan@unam.mx

\end{flushleft}
\section*{Abstract}
The entropic associative memory (EAM) is a computational model of natural memory incorporating some of its putative properties of being associative, distributed, declarative, abstractive and constructive. Previous experiments satisfactorily tested the model on structured, homogeneous and conventional data: images of manuscripts digits and letters, images of clothing, and phone representations. In this work we show that EAM appropriately stores, recognizes and retrieves complex and unconventional images of animals and vehicles. Additionally, the memory system generates meaningful retrieval association chains for such complex images. The retrieved objects can be seen as proper memories, associated recollections or products of imagination.



\section*{Introduction}
Associative memory models are a topic of intense study and development since the latter part of the previous century.  Hopfield's work \cite{hopfield_1982} is pivotal within the field of neural networks and dynamic systems, and many other models of associative memories follow that path \cite{ritter_1999, krotov_2016,danihelka_2016,he_2019,le_2020}. However, such models lack the abstractive and constructive aspects of natural memory.

The entropic associative memory (EAM) model addresses such limitations using a declarative representation in the form of 
a bi-dimensional table of $n$ columns by $m$ rows. This table, known as Associative Memory Register (AMR), stores functions taking $n$ arguments each holding one of $m$ possible values. These functions represent objects and constitute memory traces in the AMR. The memory defines three operations on an AMR and a given cue, namely memory register, memory recognition and memory retrieval, denoted as $\lambda$-register, $\eta$-recognition, and $\beta$-retrieval, respectively. 

The initial experiments \cite{pineda_2021,morales_2022} stored, recognized and retrieved manuscript digits, as well as upper and lower case manuscript letters of the corpus MNIST \cite{lecun_2010} and EMNIST \cite{cohen_2017}, respectively. The memory defined a particular AMR to store objects of each class, so there were 10 AMRs for the first experiment and 36 AMRs for the second. Moreover, the cells in an AMR were simply on and off corresponding to the argument's values of the stored functions.

The system evolved into a weighted version (Weighted Entropic Associative Memory or W-EAM) in which the memory cells have an associated weight, indicated by a natural number. So every time the $\lambda$-register operation registers a cue the value of all the involved cells increases by one, implementing a form of the Hebb's learning rule \cite{pineda_img_2023}. In this setting, every column of the table defines a probability distribution with its associated entropy. In \cite{pineda_2022}, it is shown that W-EAM can adequately store Mexican Spanish phone representations of the DIMEx100 corpus \cite{pineda_2009}. 

In the most recent experiment \cite{pineda_img_2023} all stored objects are condensed within one AMR regardless of their class. In this case the memory system stores, recognizes and retrieves images of clothing from the Fashion-MNIST data set \cite{Zalando_2017}.

In summary the properties of the 
EAM/W-EAM paradigm are,
\begin{enumerate}[label=\roman*.]
    \item Associative: all memory operations need a cue as input; this means that the model is content-based addressable.
    \item Distributed: the relation between cells and represented objects is {\em many-to-many}, i.e., an object is stored in multiple cells in the table, and a cell can be shared by multiple stored objects \cite{hinton_1986}.
    \item Declarative: the three memory operations are readily performed on the table by the execution of simple symbolic procedures without search.
    \item Abstractive: the $\lambda$-register operation performs an abstraction of the cue and the current state of the AMR.
    \item Productive: the overlapping of memory traces in the AMR gives rise to new traces besides the explicitly registered ones. This property also includes the production of false recollections.
    \item Indeterminate: the stored objects are overlapped and it is not possible to tell them apart by direct inspection.
    \item Entropic: the indeterminacy implies the entropic character of the memory; moreover, the memory follows {\em the entropy trade-off} stating that if the entropy is low the precision of the memory recognition is high but the recall is low; conversely, if the entropy is high, the recall is higher but the precision lowers; however, there is a range of moderate entropy values in which the precision and the recall are both high, and the memory has a satisfactory performance. 
    \item Direct rejection: the recognition test is defined as the logical material implication between the cue and the memory content.
    \item Constructive: the retrieval operation randomly constructs objects out of the cue's values and the probability distribution of the corresponding columns in the AMR; so it is plausible to generate memories strongly resembling the cue, recollections associated to it or even inventive objects, but also noise. 
    \item Parallel execution: the memory operations are cell to cell and column to column, and can be performed in parallel if the appropriate hardware is available (it is not the case for the current work).
    \item Large capacity: the number of objects stored at a given state is $2^{en}$, where $n$ is the number of arguments of the functions stored in an AMR with entropy value $e$ \cite{pineda_img_2023}.
    \item Low in computing resources:  the model, its operations and data structures consume low computer memory and processor resources.
    \item Autoassociative and to certain degree heteroassociative:  the autoassociative property is intrinsic to the design of the model, whereas a sort of heteroassociative behavior is a side effect of the memory operations. 
\end{enumerate}

The properties of EAM/W-EAM make it a substantially different model from the Neural Network approaches of associative memory, as has been extensively discussed in \cite{pineda_2021,morales_2022,pineda_2022,pineda_img_2023}. 

The experiments above use structured and homogeneous data to test EAM/W-EAM, such as conventional images of manuscript digits and letters, highly stereotyped images of clothing, and coefficients (MFCCs) representing features of Mexican Spanish phones. In this paper we investigate the performance of W-EAM with only one AMR, as done in \cite{pineda_img_2023}, but storing more complex data than before, i.e., images of animals and vehicles from the real world contained in the \Cifar\cite{Krizhevsky_2009} corpus.

\section*{Related work}
Computational models of associative memory have been traditionally implemented using artificial neural networks (ANN); being Hopfield's \cite{hopfield_1982} the paradigmatic model. Lately, several brain-inspired models of associative memories have been developed with ANN \cite{he_2019, kozachkov_2023, zhang_2023, jimnez_2023}. Although the neural network paradigm simulates the biological functionality of the human memory, most ANNs are trained, store the inputs separately, retrieve the stored objects exactly avoiding additional o spurious values \cite{palm_1980,willshaw_1969,simas_2023}, and never reject unless a pre-established threshold is exceeded, which oppose the characteristics of EAM/W-EAM.

The present model stores functions corresponding to abstract amodal representations of the modality-specific input objects, as discussed below. The approaches in \cite{simas_2023,annabi_2022,kozachkov_2023} also preprocessed the data before store it.

In other models the retrieval operation has a probabilistic aspect in terms of the strengths between cues and memory objects \cite{Raaijmakers_1980}, and of the posterior distribution of each initial pattern to reproduce the cue \cite{MacKay_1991,Sommer_1998}. However, the retrieved objects are photographic reproductions of the cues. The model in \cite{Liu_2019} reconstructs the retrieved object after several iterations aimed at maximizing a probability distribution with respect to the cue. The approach in \cite{Alvesdasilva_1992} models interdependencies of events and of restricted variables to estimate the values of the retrieved object; however, it does not require a cue and can be considered a free recall. Another approach \cite{janusz_2007} uses a hierarchical array of processing elements storing probabilities to compute an output function, so a feedback operation makes associations and recovers missing parts of the input data.  

There are also approaches that use the entropy in specific aspects of the memory retrieval operation \cite{Sommer_1998, Nakagawa_2006}, and also knowledge acquisition \cite{Alvesdasilva_1992}; however, this narrow use of the entropy contrasts with its relevance in the overall performance, capacity and operation of EAM/W-EAM.

The study of associative memories storing and retrieving images from the real word have been addressed with ANNs, such as: a deep associative neural network with a probabilistic learning method similar to the Restricted Boltzmann Machine \cite{Liu_2019}; a combination of predictive coding and a generative network \cite{salvatori_2021}; a bidirectional associative memory incorporating hidden layers and supervised backpropagation \cite{kosko_2021}; a variational autoencoder providing estimates and predictions to Gaussian Mixtures Models \cite{annabi_2022}; and a neuron-astrocyte network \cite{kozachkov_2023}. Our present proposal contrasts radically with such approaches, and the experiments provide  satisfactory results suggesting that our overall framework is very promising.

\section*{Functionality and architecture of the system}
In this section we present the functional specification and architecture of the memory system. We formally define an AMR of size $n\times m$ in W-EAM as the weighted relation $r:A\mapsto V$, where $A$ holds the set of arguments $\{a_1,\ldots,a_n\}$, and $V$ holds the set of values $\{v_1,\ldots,v_m\}$. The function $R:A\times V\mapsto W$ specifies the weight of the tuples in $r$, where $W$ holds a finite set of natural numbers, such that $R(a_i,v_j)=w_{ij}>0$ if and only if $(a_i,v_j)\in r$.

Next, we introduce the functionality of the memory by describing the actions of the system to perform the memory operations. All three operations have in common the first four actions. (1) A concrete object cues the memory. (2) The system encodes this object as a real-valued function of $n$ arguments. (3) The quantization maps these arguments into one of $m$ integer values yielding the function $f_a:A\mapsto V$. (4) The system carries out the corresponding memory operation on:

\begin{itemize}
    \item The AMR of size $n\times m$ storing all the data registered in the memory, formalized as the weighted relation $r_f:A\mapsto V$.
    \item The quantized function $f_a$ shaped into an auxiliary AMR of size $n\times m$, formalized as the weighted relation $r_a:A\mapsto V$, such that $(a_i,v_j)\in r_a$ if and only if $f_a(a_i)=v_j$. Additionally,  $R_a(a_i,v_j)=1$ for all $(a_i,v_j)\in r_a$.
\end{itemize}

\begin{figure}[htbp]
\centering
\includegraphics[width=1\textwidth]{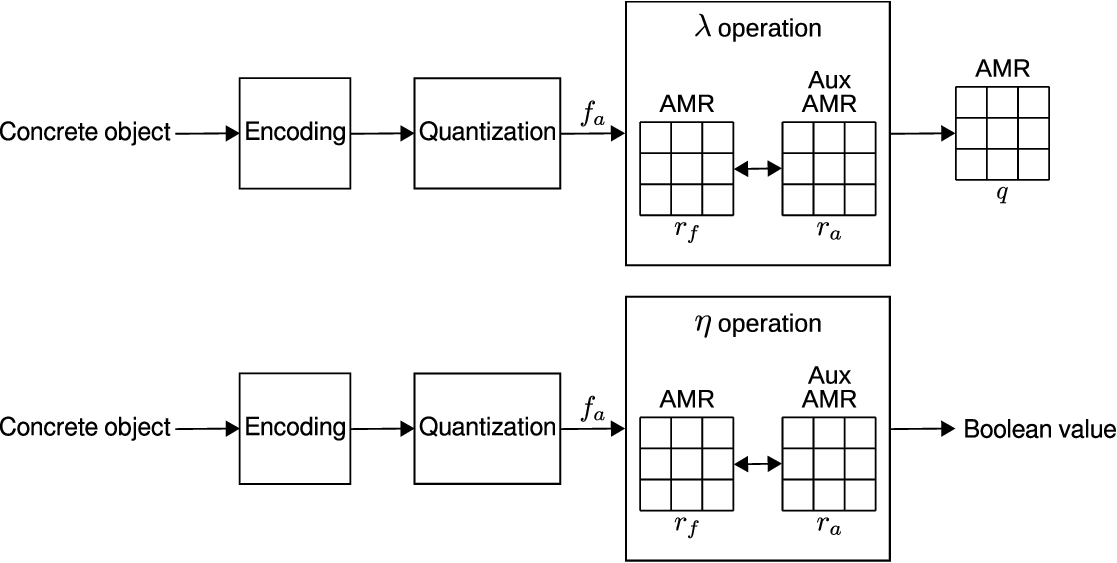}
\caption{\bf{Actions taking place in the $\lambda$-register (top) and $\eta$-recognition (bottom) operations.}}
\label{fig:f1-lambda-eta-operation}
\end{figure}

The $\lambda$-register operation (see top of Fig. \ref{fig:f1-lambda-eta-operation}) generates a new state of the AMR, that we denote $q$, by incorporating $r_a$ in $r_f$.

The $\eta$-recognition operation (see bottom of Fig. \ref{fig:f1-lambda-eta-operation}) produces a boolean value indicating whether $r_a$ is contained in $r_f$. 

\begin{figure}[htbp]
\centering
\includegraphics[width=0.8\textwidth]{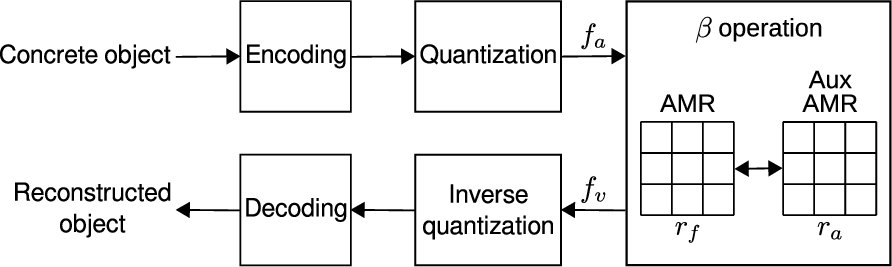}
\caption{\bf{Actions taking place in the $\beta$-retrieval operation.}}
\label{fig:f2-beta-operation}
\end{figure}

The $\beta$-retrieval operation (see Fig. \ref{fig:f2-beta-operation}) constructs the function $f_v:A\mapsto V$. An inverse quantization computes  real-valued features for $f_v$. The system decodes these features into a concrete representation. This reconstructed object is the retrieval of the memory. 

The system defines three representational levels for the data involved in the operations.

\begin{enumerate}
    \item Modality-specific concrete representations corresponding to the concrete objects that cue the memory in Figs. \ref{fig:f1-lambda-eta-operation} and \ref{fig:f2-beta-operation}, which are images of the \Cifar\ corpus, with or without noise. The \Cifar\ corpus consists of 60,000 images of $32\times 32$ RGB pixels labeled as one of ten classes in Table \ref{tab:tab1-Cifar10-classes}, with 6,000 instances of each class. The size of a \Cifar\ image is $pixel\ width\times pixel\ height \times channels = 32\times 32\times 3$ $=3,072$. 

    \begin{table}[htbp]
    \centering
    \begin{tabular}{| m{4em} | l | m{4em} | l |}
    \hline
    \rowcolor{Gray}
    \bf{Class number} & \bf{Class name} & \bf{Class number} & \bf{Class name}\\
    \hline
    0  & Airplane & 5 & Dog \\
    \hline
    1  & Automobile & 6 & Frog \\
    \hline
    2  & Bird & 7 & Horse \\
    \hline
    3  & Cat & 8 & Ship \\
    \hline
    4  & Deer & 9 & Truck \\
    \hline
\end{tabular}
\caption{{\bf Classes of the} \Cifar\ {\bf corpus.}}
\label{tab:tab1-Cifar10-classes}
\end{table}
    \item Abstract amodal representations, which include:
    \begin{itemize}
        \item The  real-valued functions encoding the concrete objects in the previous level. These functions result of the encoding component in Figs. \ref{fig:f1-lambda-eta-operation} and \ref{fig:f2-beta-operation}.
        \item The function that the inverse quantization computes, as in Fig. \ref{fig:f2-beta-operation} .
    \end{itemize}  

    \item Abstract amodal distributed  representations consisting of the quantized functions held in the AMR. These functions are:
    \begin{itemize}
        \item $f_a$ in Fig. \ref{fig:f1-lambda-eta-operation}, which  the $\lambda$-register operation stores  and the $\eta$-recognition operation recognizes in the AMR.
        \item $f_a$ and $f_v$ in Fig. \ref{fig:f2-beta-operation}, which the $\beta$-retrieval operation uses as cue and retrieves from the AMR, respectively.
    \end{itemize}
    
\end{enumerate}

Since the $\beta$-retrieval operation involves more parts of the system than the other two operations, we examine in detail the components of W-EAM in the order presented in Fig. \ref{fig:f2-beta-operation} from left to right, pairing complementing parts if available. Thus the components are: encoder and decoder, quantizier and its inverse, and memory operations.

\subsection*{Encoder and decoder}
We use an encoder and a decoder as the scaffolding to map concrete objects into and from abstract amodal representations, respectively. We take an autoencoder \cite{hinton_2006,masci_2011} to implement the encoder and decoder pair, which is a neural network with two parts: (1) an encoder that performs dimensionality reduction, and (2) a decoder that  tries to reproduce the initial data from the output of (1). The experiments below examine the behavior of the autoencoder and the memory model when reducing the 3,072 values of the \Cifar\ images to a latent space of size $n\in\{2^i\,|\, 6\leq i \leq 10\}$. 

Furthermore, the EAM uses a classifier of the \Cifar\ data set whose training occurs on the latent space. In the basic model  \cite{pineda_2021,morales_2022}, this classifier was only an aid to the training of the encoder. But later \cite{pineda_2022,pineda_img_2023} it also became a tool for the improvement and development of the system. The simultaneous training of the encoder and the classifier negatively affects the performance of the autoencoder for \Cifar, despite producing satisfactory results in previous experiments. Hence, we first train the entire autoencoder, and then the classifier. 

We design the autoencoder and the classifier after the neural networks in \cite{Fagbohungbe_2022}. The former includes convolutional, max pooling and upsampling layers, as well as ReLU as the activation function and root mean squared error (RMSE) as the metric. The latter contains convolutional, max pooling, dropout and fully connected layers, as well as ReLU as the activation function and accuracy as the metric. Moreover, to reduce overfitting in the classification we carry out data augmentation by horizontal flipping, minor rotations and slight zooming on the training images. We inserted in the classifier two additional elements to the  architecture in \cite{Fagbohungbe_2022}, namely, batch normalization layers and weight decay ---a parameter that prevents overfitting \cite{giuste_2020}. 

The training of the autoencoder and the classifier, and the execution of the experiments below, implement a 10-fold cross-validation procedure. Each fold considers a partition of the corpus into three disjoint sets as follows, 

    \begin{itemize}
        \item \emph{Training Corpus} for training the autoencoder and the classifier: 70\%.
        \item \emph{Remembered Corpus} for filling an AMR used during the execution of the experiments: 20\%.
        \item \emph{Test Corpus} for testing the networks and the performance of the memory system: 10\%.
    \end{itemize}

In turn the \emph{Training Corpus} is split into two parts: for training the networks 80\%, and for validating the results 20\%.
 
\begin{figure}[htbp]
\centering
\includegraphics[width=1\textwidth]{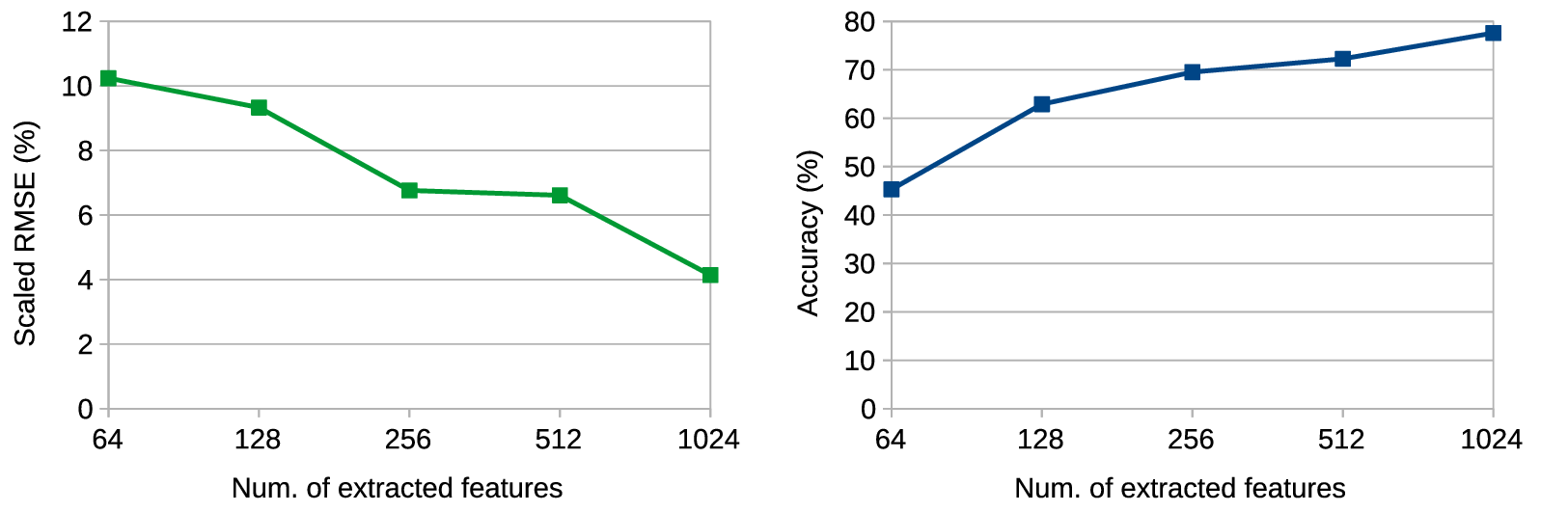}
\caption{{\bf Performance of the networks.} The left side presents the mean RMSE of the autoencoder, this error is scaled to a percentage of the maximum value (255) for an RGB channel. The right side displays the mean accuracy values of the classifier. }
\label{fig:f3-models-performance}
\end{figure}

Fig. \ref{fig:f3-models-performance} shows the performance of the current networks, which increases as more features are extracted. The scaled RMSE of our autoencoder is small compared with the stated in \cite{Fagbohungbe_2022}, where the autoencoder for all number of features has a MSE of at least 0.54, corresponding to a scaled RMSE of 73.48. The accuracy of our classifier is slightly higher than in \cite{Fagbohungbe_2022}. For example, we obtain an accuracy of 72.27\% for 512 extracted features; whereas \cite{Fagbohungbe_2022} reports an accuracy value less than 70\% for 768 features. The state-of-the-art classifiers for \Cifar\ achieve nearly 100\% accuracy \cite{dosovitskiy_2021}, with pre-trained visual transformer models on large (and curated \cite{oquab_2023}) amounts of data. However, such classifiers work on concrete data, as opposed to a  latent space and cannot be used in the present setting. 

\subsection*{Quantizer and its inverse}
The quantizer maps the $n$ real-valued features into $m$ discrete levels as follows (see Eq. \ref{eq:new_quantizer}):

\noindent For $1\leq i\leq n$,
\begin{enumerate}[label=\Roman*.]
    \item Perform the {\em min-max} normalization on the $i$-th feature of the images cueing the memory. The normalization draws the minimum and maximum values from the $i$-th feature of the cues in the {\em Remembered Corpus}, which are denoted $f_i^{min}$ and $f_i^{max}$, respectively. 
    
    So, the normalization of the $i$-th feature, named $feature_i$, is:
    \[g_i=\displaystyle\frac{feature_i-f_i^{min}}{f_i^{max}-f_i^{min}}.\]
    \item Find the row between 0 and $m-1$ corresponding to $g_i$.
\end{enumerate}
\begin{equation}\label{eq:new_quantizer}
    quantizer(feature_i, f_i^{min}, f_i^{max}, m) =             round\left(g_i\cdot(m-1)\right)
\end{equation}

For the inverse quantizer we need to avoid a division by zero  when $m=1$. In such case, the inverse quantizer is half the distance between $f_i^{min}$ and $f_i^{max}$ (see Eq. \ref{eq:new_inverse_quantizer}).
\begin{equation}\label{eq:new_inverse_quantizer}
   \begin{split}
    inv\_quantizer(&quantized_i, f_i^{min}, f_i^{max}, m) = \\ 
        &\begin{cases}
        \dfrac{f_i^{max}-f_i^{min}}{2} & \text{if\ } m=1\\[0.9em]
        f_i^{min} + \left(\dfrac{quantized_i\cdot(f_i^{max}-f_i^{min})}{m-1} \right) &\text{otherwise}
        \end{cases}
    \end{split}
\end{equation}

\subsection*{Memory operations}
The definitions of the memory operations \cite{pineda_img_2023,pineda_2024} are as follows: 

\begin{enumerate}
    \item {\em Memory register} (see top of Fig. \ref{fig:f1-lambda-eta-operation}).
    
    $\lambda(r_f,r_a)=q$, such that $q=r_f\cup r_a$. So, $Q(a_i, v_j)=R_f(a_i,v_j)+R_a(a_i,v_j)$ for all $a_i \in A$ and $v_j\in V$ ---i.e., the system computes the weights of the new state $q$ adding the corresponding weights of the previous state $r_f$ and of the cue $r_a$. Hence, $Q(a_i, v_j)=R_f(a_i,v_j)+1$ if $(a_i,v_j)\in r_a$, and $Q(a_i, v_j)=R_f(a_i,v_j)$ otherwise. 
    
    Initially, $r_f$ is the empty relation; consequently, $R_f(a_i,v_j)=0$ for all $(a_i,v_j)\in A\times V$.
    \item {\em Memory recognition} (see bottom of Fig. \ref{fig:f1-lambda-eta-operation}).
    
    $\eta(r_a,r_f,\iota,\kappa,\xi)$ is the boolean value corresponding to $\big(R_a(a_i,v_j)>0\rightarrow R_f(a_i,v_j)\geq\iota \omega_i\big) \wedge \rho\geq\kappa\Omega$, such that the implication holds for all values $v_j$ of at least $n-\xi$ arguments $a_i$ of $r_f$ (i.e., material implication relaxed by $\xi$). Additionally,

    \begin{enumerate}
        \item $\omega_i=\dfrac{1}{k}\displaystyle\sum^m_{j=1}R_f(a_i,v_j)$, where $k$ is the number of cells in the column $i$ of $r_f$ with $\omega_{ij}\neq 0$; i.e., the average weight of argument $a_i$ among its values $v_j$ with non-zero weight.
        \item $\Omega=\dfrac{1}{n}\displaystyle\sum^n_{i=1}\omega_i$; the average weight for all $\omega_i$.
        \item $\rho=\dfrac{1}{n}\displaystyle\sum^n_{i=1}R_f(a_i,v_{cue})$, where $v_{cue}\in V$ is the only value for which $R_a(a_i,v_{cue})>0$; the weight of the cue.
    \end{enumerate}
    The parameter $\iota$ in $R_f(a_i,v_j)\geq\iota \omega_i$ modulates whether the cell $(a_i,v_j)\in r_f$ is on or off; and the parameter $\kappa$ in $\rho\geq\kappa\Omega$ controls the strength of the cue to be accepted.

    \item {\em Memory retrieval} (see Fig. \ref{fig:f2-beta-operation}).
    
    $\beta(f_a,r_f,\sigma)=f_v$, such that if $\eta(r_a,r_f,\iota,\kappa,\xi)$ holds, then for each argument $a_i$ the system randomly assigns $f_v(a_i)$ a value $v_j$ from $r_f(a_i)$ ---i.e., the column $i$--- as follows:

    \begin{enumerate}
        \item Let $\zeta_i$ be a normal distribution centered at $f_a(a_i)$ with standard deviation $\sigma m$.         
        \item Let $\Psi_i$ be the probability distribution formed by the weights in the column $i$.
        \item Let $\Phi_i$ be the product of $\Psi_i$ and $\zeta_i$.
        \item $f_v(a_i)$ is randomly selected from $\Phi_i$.
    \end{enumerate}

    If $\eta(r_a,r_f,\iota,\kappa,\xi)$ does not hold, then $\beta(f_a,r_f,\sigma)$ is undefined ---i.e., $f_v(a_i)$ is undefined for all $a_i$.

    The parameter $\sigma\geq 0$ measures the similarity of the reconstructed object with respect to the cue. The $\beta$-retrieval operation can produce objects identical, related and distant to the cue, or even noise, according to the value of $\sigma$.
\end{enumerate}

The entropy of column $i$ is Shannon's entropy of the probability distribution $\Psi_i$, and the entropy of the memory is the average entropy of the columns in the AMR.

\begin{figure}[htbp]
\centering
\includegraphics[width=1\textwidth]{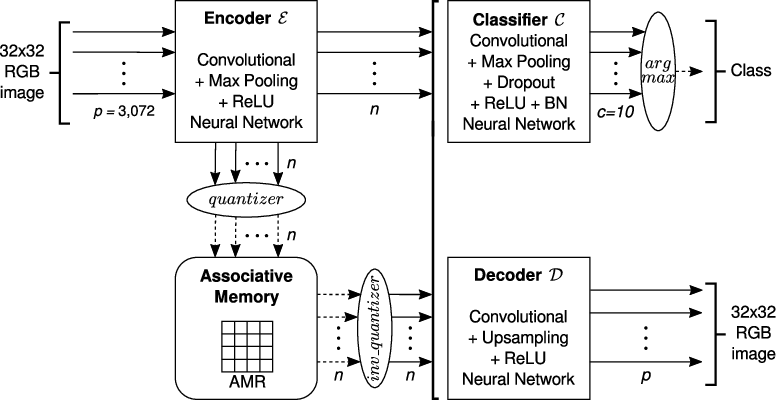}
\caption{{\bf System architecture.} The variable $p$ denotes the size of the \Cifar\ images, $n$ indicates the number of extracted features, and $c$ is the size of the probability distribution of classifying a cue in the latent space as one the \Cifar\ classes. Out of such distribution $argmax$ chooses the class number with the highest probability. Solid and dashed lines represent real and integer numbers, respectively. }
\label{fig:f4-architecture}
\end{figure}

Fig. \ref{fig:f4-architecture} depicts the architecture of the W-EAM system with one AMR on the \Cifar\ data set. \textsf{BN} in the classifier's box refers to batch normalization.  $\mathcal{E}$, $\mathcal{D}$ and $\mathcal{C}$ represent the encoder, decoder and classifier, as shown in the corresponding boxes.

\section*{Experiments and results}
We conducted five experiments analogous to \cite{pineda_img_2023} as follows:

\begin{enumerate}
    \item Determine the optimal size and filling percent of the AMR by examining the performance of the system under different values of $n$ and $m$, as well as various portions of the \emph{Remembered Corpus}. 
    \item Analyze the performance of the $\beta$-retrieval operation by taking different values of the parameter $\sigma$. For each $\sigma$ we show instances of the retrieved objects of the ten classes of \Cifar.
    \item Run experiment 2 on the same settings but with noisy cues.
    \item Generate association chains originating from complete cues.
    \item Run experiment 4 but using noisy cues.
\end{enumerate}

\subsection*{Experiment 1}
We determine the size $n\times m$ of the best AMR filled up with the whole of the {\em Remembered Corpus} and test the system on the {\em Test Corpus},  where $n\in\{2^i\,|\,6\leq i\leq 10\}$ and $m\in\{2^j\,|\,0\leq j\leq 9\}$.

The metrics that we use to measure the memory performance are precision, recall, accuracy and F1 score, defined in the standard way. The AMR is unable to classify the stored data, so $\mathcal{C}$ predicts the class of the retrieved objects, which we compare with the  actual classes of the cues. A true positive occurs when the class of the cue and of the retrieved object is the same. Otherwise, we have a false positive for the class $\mathcal{C}$ predicts, and also a false negative for the actual class of the cue. In case the AMR rejects the cue, there is a false negative for the system as a whole because all objects in the {\em Test Corpus} belong to one of the classes. Furthermore, there are no true negatives and the system counts every false positive as a false negative too, so the accuracy and the recall are the same.

The higher the values of the parameters $\iota$ and $\kappa$ the harder the AMR accepts the cue, decreasing false positives but increasing false negatives; conversely, the parameter $\xi$ relaxes the $\eta$-recognition operation preventing false negatives. The impact of these parameters is investigated in \cite{pineda_2022}, although it is not the focus of the present experiments and we use their default values, which is zero in all three cases. Likewise, we set $\sigma=0.025$, this small value offers a good compromise between the performance of the system and the reconstruction of objects not only similar to the cue, but also associated or imaged recollections. In particular, the retrievals are most likely the exact (quantized) cues for $m<32$. Moreover, slight increments of $\sigma$ negatively affect the memory performance. Thus, precise memories are hard to reconstruct and the system is highly sensitive to the parameter $\sigma$. 

\begin{figure}[htbp]
\centering
\includegraphics[width=0.85\textwidth]{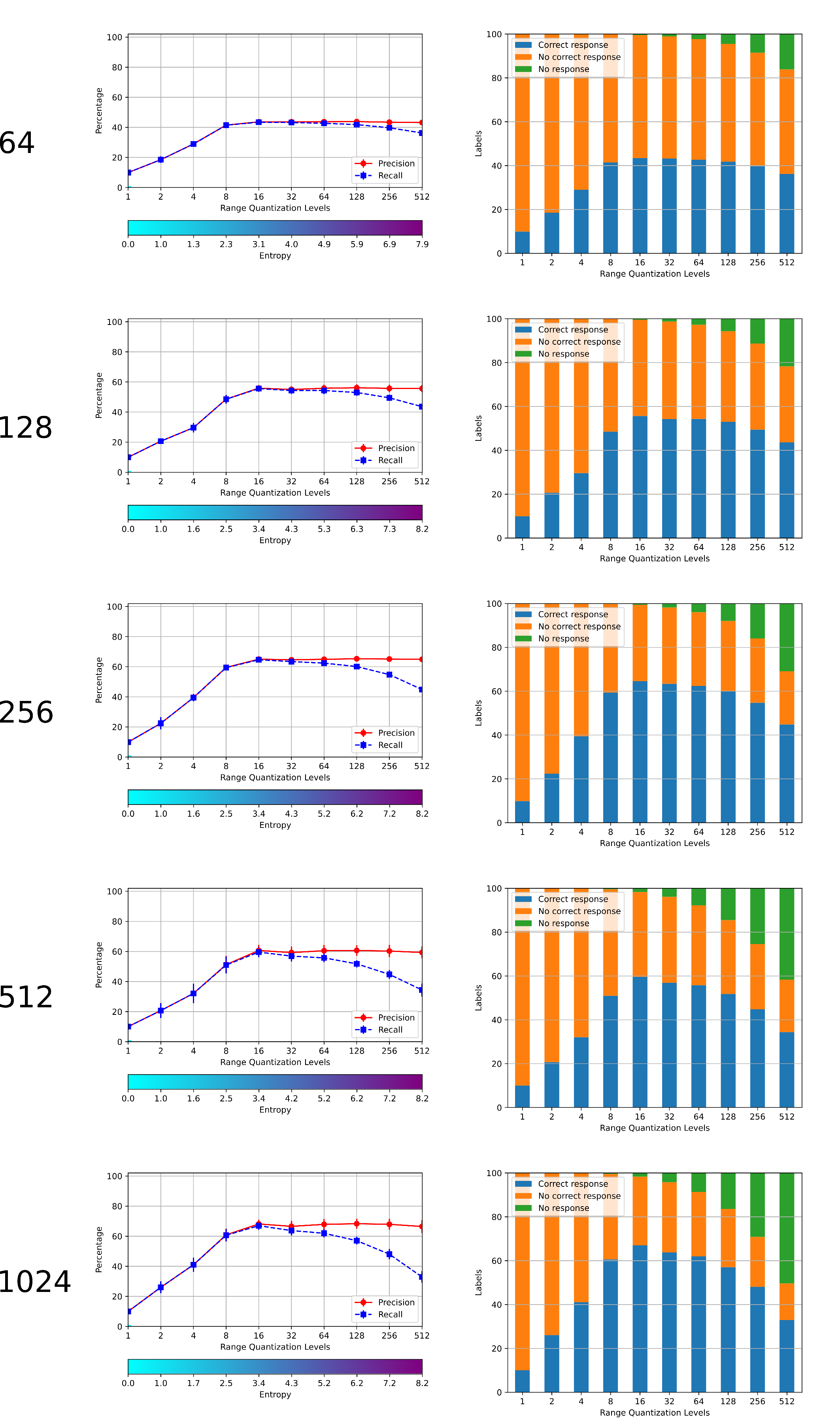}
\caption{{\bf Memory performance for the AMRs of size $n\times m$.} The large numbers located at the left indicate the number of columns. The \textsf{Range Quantization Levels} in the $X$-axes correspond to the number of rows. The line graphs show the precision and the recall for each AMR, with the associated entropy values at the bottom horizontal bar. The bar charts display the type and amount of responses for each register.}
\label{fig:f5-amr-performance}
\end{figure}

Fig. \ref{fig:f5-amr-performance} shows the precision, the recall and the entropy values, as well as the type of response for each register of size $n\times m$.  The highest accuracy of the memory system is 66.97\% for the AMR of size $1{,}024\times 16$. The second highest accuracy is 64.63\% for the AMR of size $256\times 16$. Moreover, the retrievals of the former are closer to the cues than those of the latter. So, we use the register with 1,024 columns in subsequent experiments. 

\begin{figure}[!htbp]
\centering
\includegraphics[width=\textwidth]{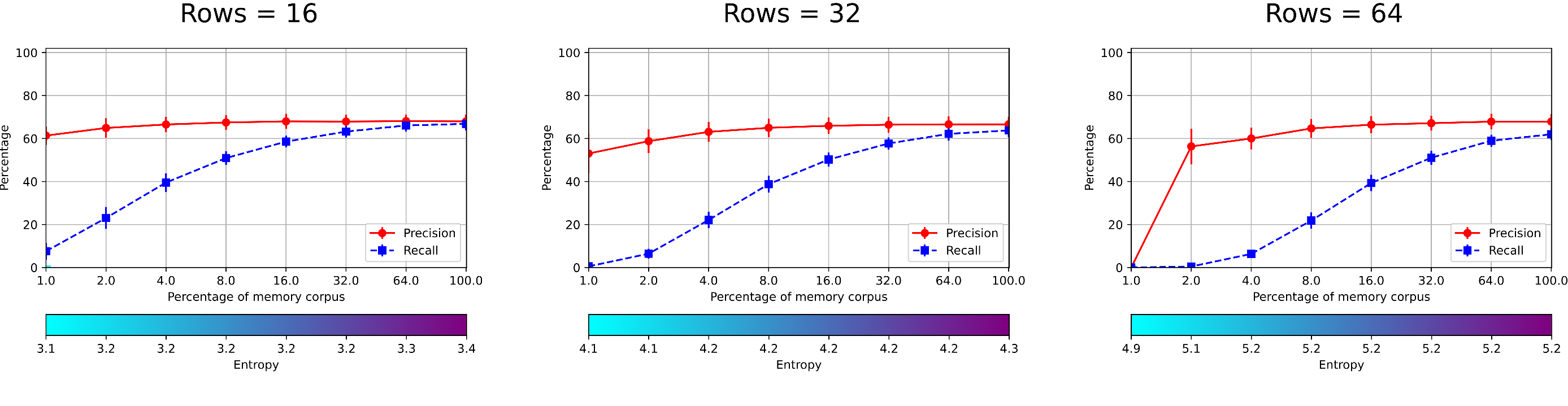}
\caption{{\bf Performance of the memory filled with portions of the {\em Remembered Corpus}.}}
\label{fig:f6-performance-AMR-1024-columns}
\end{figure}

We study the behavior of such register with 16, 32 and 64 rows because of their high F1 scores. The system fills these AMRs with portions of 1\%, 2\%, 4\%, 8\%, 16\%, 32\%, 64\% and 100\% of the \emph{Remembered Corpus}. Fig. \ref{fig:f6-performance-AMR-1024-columns} shows for each filling percentage the precision, the recall and the entropy of the AMRs tested on the \emph{Test Corpus}. The memories require a large amount of the \emph{Remembered Corpus} to operate, with their best results at 100\%. We pick the AMR with 16 rows for its smaller size and higher performance. Therefore, the experiments below used an AMR of size $1{,}024\times 16$.

Figs. \ref{fig:f5-amr-performance} and \ref{fig:f6-performance-AMR-1024-columns} corroborate the entropy trade-off. In particular, the AMR of size $1{,}024\times 64$ in Fig. \ref{fig:f6-performance-AMR-1024-columns} becomes operative at its highest entropy, which is a moderate value considering all entropies of the AMRs with 1,024 columns.

\begin{figure}[htbp]
\centering
\includegraphics[width=\textwidth]{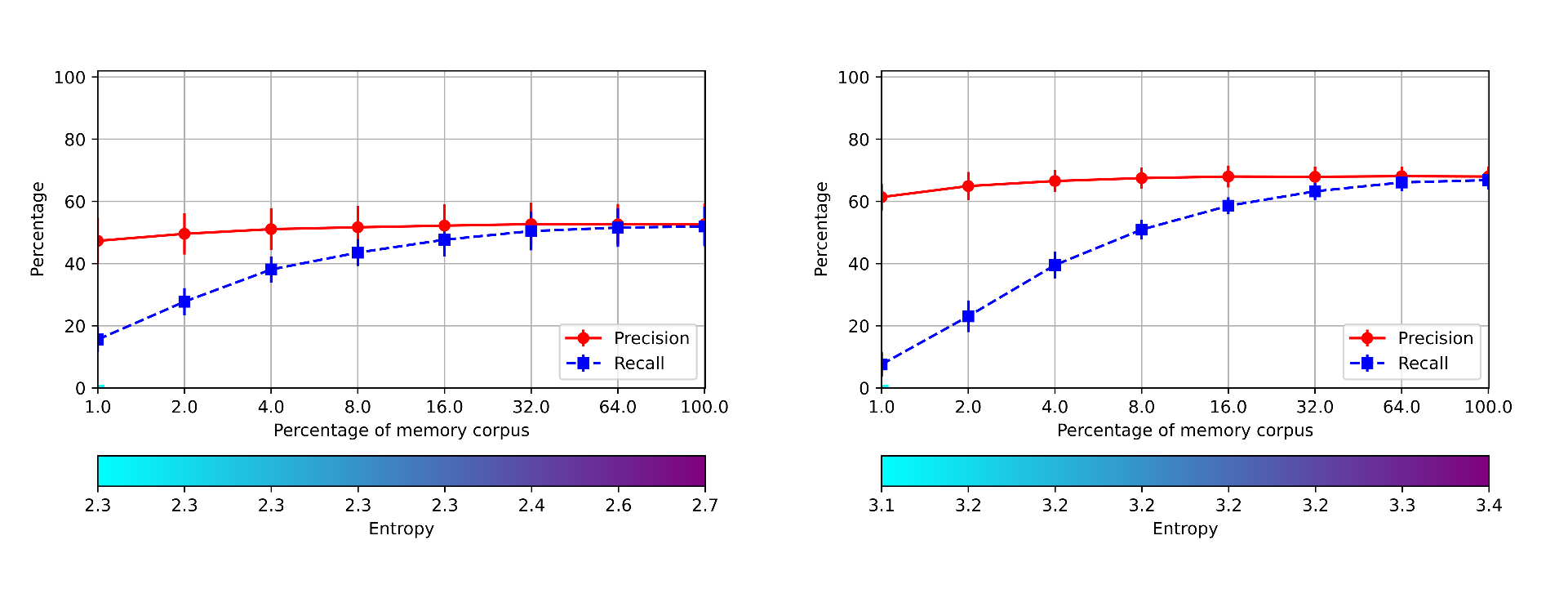}
\caption{{\bf Effect of the quantizer on the memory performance.} The left and right side show for each filling percentage the performance of the AMR of size $1{,}024\times 16$ using the previous and current version of the quantizer, respectively.}
\label{fig:f7-quantizers-performance}
\end{figure}

In previous works, the normalization in the quantizer was a general process involving all features of the cues. In the current investigation the normalization takes place locally for each feature, improving significantly the performance of the system, as shown in Fig \ref{fig:f7-quantizers-performance}.

\subsection*{Experiment 2}
\begin{figure}[!htbp]
\centering
\includegraphics[width=\textwidth]{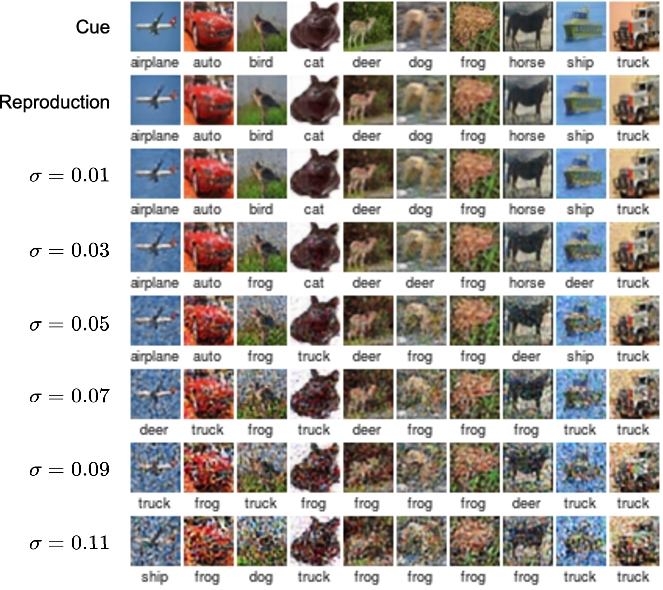}
\caption{{\bf Examples of reconstructions by the $\beta$-retrieval operation.}}
\label{fig:f8-experiment-2}
\end{figure}

We test the reconstructions of the $\beta$-retrieval operation for six values of the parameter $\sigma$ uniformly spread along the range 0.01 to 0.11, as shown in Fig. \ref{fig:f8-experiment-2}. The first row exhibits ten cues from each class of \Cifar. The second row shows the output of the autoencoder directly, without using the memory; as can be seen the class of such output is correct. The third through the eighth rows show the retrieved images for the corresponding values of $\sigma$. The label below each image in the first row indicates the class of the input image, and the labels in the remaining rows represent the class produced by $\mathcal{C}$ of the respective image.

We identify four types of reconstructions in relation to the image quality and the class in the label:
\begin{itemize}
    \item {\em Remembered object} has a good quality and is assigned the correct class.
    \item {\em Associated object} has a good quality, although a different class.
    \item {\em Imaged object} has a lower quality, yet can be interpreted regardless of its class. 
    \item {\em Noise} otherwise. 
\end{itemize}

For the $\sigma$ values of 0.01 and 0.03, the reconstructions are remembered objects, with three exceptions under $\sigma=0.03$: the bird is classified as a frog, the dog as a deer and the ship also as a deer, which we consider associations. The image quality for $\sigma=0.05$ deteriorates but is still satisfactory so the reconstructions are remembered objects in six cases: airplane, automobile, deer, frog, ship and truck; and associations in the other instances: the bird is classified as a frog, the cat as a truck, the dog as a frog, and the horse as a deer. The quality of the retrieved images for $\sigma>0.05$ considerably declines  obtaining imaged objects or noise. The frog and the truck for $\sigma>0.05$, and the deer for $\sigma=0.07$ are imaged objects, as well as the automobile interpreted as a truck ($\sigma=0.07$), the bird as a dog ($\sigma=0.11$), the cat as a truck ($\sigma\in\{0.07,0.11\}$), the horse as a deer ($\sigma=0.09$), and the ship as truck ($\sigma\in\{0.07,0.09\})$. The rest of the retrieved images cannot be interpreted and are considered noise.

\subsection*{Experiment 3}

\begin{figure}[htbp]
\centering
\includegraphics[width=\textwidth]{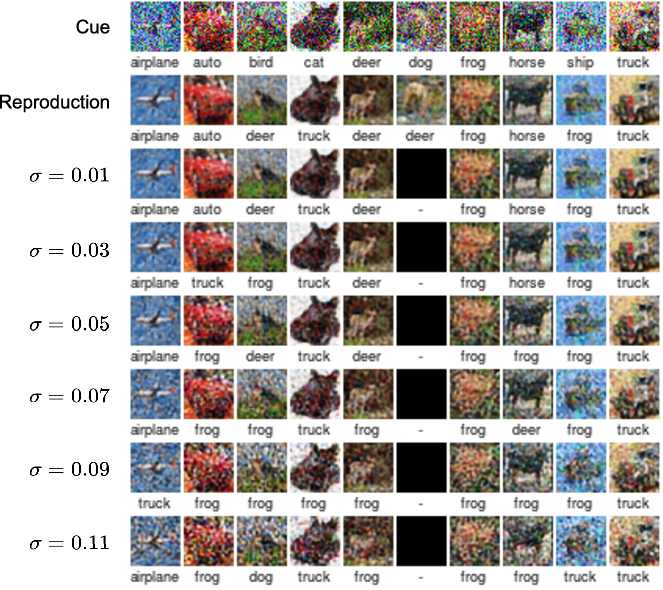}
\caption{{\bf Examples of reconstructions by the $\beta$-retrieval operation with noisy cues.}}
\label{fig:f9-experiment-3}
\end{figure}

This experiment is analogous to experiment 2 but using noisy inputs.  The retrieved objects are shown in Fig. \ref{fig:f9-experiment-3}. 
The noisy input cues in the first row are severe distortions of their source images. Nonetheless, the reproductions of the autoencoder in the second row provide good approximations to the original instances, however such reproductions are not as fine as the ones achieved in experiment 2, and the bird, the cat, the dog and the ship are assigned a different class. Black squares indicate that the cue is rejected as in the dog column for all values of $\sigma$. In any other respect, the retrieved images are similar to those in experiment 2 although the remembered objects are fewer, and the associations, the imaged objects and noise emerge with smaller values of $\sigma$, so most figures are quite noisy.

\subsection*{Experiment 4}
\begin{figure}[htbp]
\centering
\includegraphics[width=\textwidth]{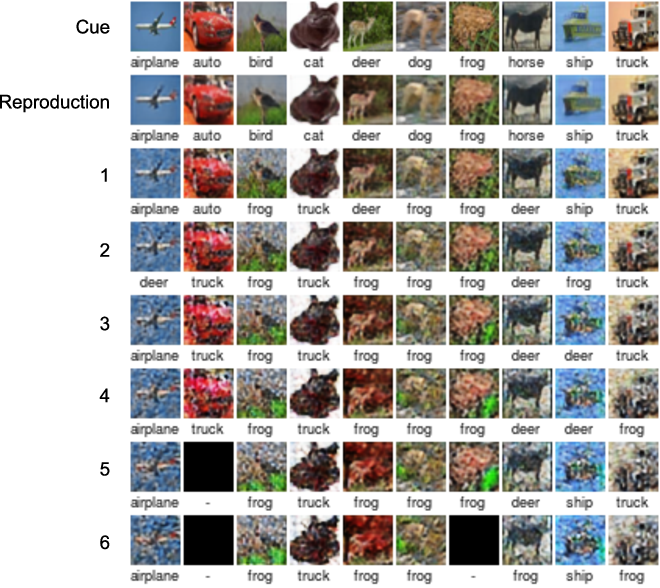}
\caption{{\bf Association chains.} The number at the left indicates the current depth level in the chain, and the name below the image is the class assigned by $\mathcal{C}$.}
\label{fig:f10-experiment-4}
\end{figure}

We obtained association chains through the $\beta$-retrieval operation with the parameter $\sigma$ set to 0.04. Fig. \ref{fig:f10-experiment-4} shows ten association chains of length six. The first row exhibits the same cues considered above from the classes of \Cifar. The second row displays the reproductions of the autoencoder for such cues. The row labeled 1 shows the objects retrieved from the original cues, and the rows labeled 2 to 6 depict the objects retrieved using the image in the previous row as the cue. 

The retrieved image quality declines as we go deeper in the association chain, such that:

\begin{itemize}
    \item Levels 1 and 2 produce remembered objects (e.g., the frog and the truck), and associations (e.g., the cat associated to the truck, and the horse to the deer). 
    \item Levels 3 through 5 produce imaged objects (e.g., the automobile interpreted as a truck, and the horse as a deer) and noise (e.g. the bird and the deer). 
    \item The objects retrieved in level 6 are pure noise.
\end{itemize}

\subsection*{Experiment 5}
\begin{figure}[htbp]
\centering
\includegraphics[width=\textwidth]{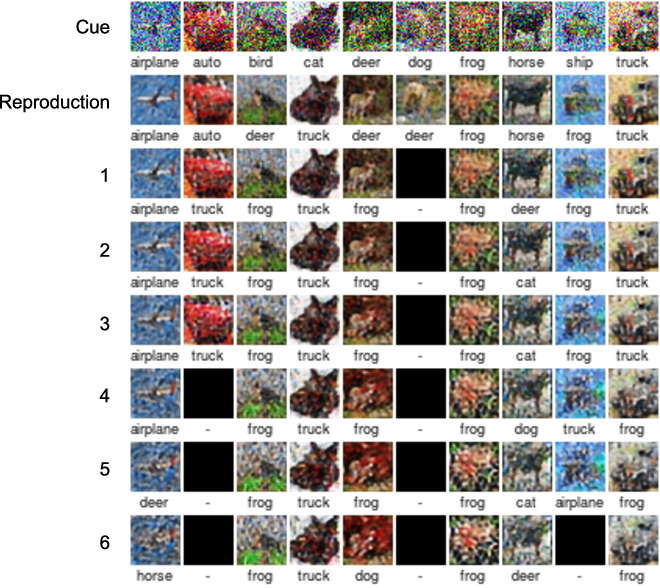}
\caption{{\bf Association chains considering noisy cues.}}
\label{fig:f11-experiment-5}
\end{figure}

This experiment is analogous to experiment 4 but with noisy cues, as shown in Fig. \ref{fig:f11-experiment-5}. The association chains are similar for cues with or without noise, although in the present experiment there are less remembered objects and associations, and more imaged objects and noise.

\subsection*{Experimental settings}
We ran the experiments using a computer with an Intel(R) Core(TM) i7-6700 CPU, 64GB of RAM and a GeForce GTX TITAN X GPU. The code is available at \url{https://github.com/nohernan/W-EAM_Cifar10/}, it is written for Python 3.10.9 and TensorFlow 2.8.3.

\section*{Discussion}
We have successfully stored, recognized and recovered images of animals and vehicles in the W-EAM system using only one AMR. This is an improvement on the kind of data used in previous experiments involving conventional representations: manuscript digits and letters, and pieces of clothing of the MNIST, EMNIST and Fashion-MNIST data sets, respectively, each consisting of structured images with 784 features; and MFCCs capturing phones of the DIMEx100 corpus in 208 features that model the human speech acoustic data \cite{Rishiraj_2013}. The images of animals and vehicles of \Cifar\ consists of 3,072 features describing different colors, shapes, backgrounds and positions. So, these images are less structured and more complex than previous domains. 

The memory system produced retrieved objects that are not alike other images in the data set, as was the case for the images of digits, letters and clothes. The retrieved objects were reconstructions of the original cues with an increasing amount of noise as $\sigma$ grew or the association chain was traversed down. Next, we describe the different outcomes of the retrieval operation:
\begin{itemize}
    \item Remember objects, which occurred for values of $\sigma$ smaller than those used for structured domains. This indicates that remembering unstructured images is harder than structured data \cite{Padgett_1987,Garland_1991,Gobet_1996,Buckley_2018}. 
    \item Objects associated to or interpreted as a class sharing some characteristics with the cue. This kind of association or interpretation is coherent and expected. Such is the case of the automobile classified as a truck, the horse as a deer, and the ship as a truck in experiments 2 and 3. Similarly, association chains show a close relationship among the cues and retrieved objects; for example the chains: automobile$\to$truck$\to$truck$\to$truck, and deer$\to$cat$\to$cat$\to$dog$\to$cat$\to$deer, as shown in Figs. \ref{fig:f10-experiment-4} and \ref{fig:f11-experiment-5}, respectively. 
    \item Objects associated to or interpreted as an unrelated class. For instance, the reconstruction of the bird in Fig. \ref{fig:f8-experiment-2} with $\sigma=0.11$ is classified as a dog, as illustrated in Fig. \ref{fig:f12-imaged_objects} (a). Another example is the retrieval of the cat in Fig. \ref{fig:f10-experiment-4} at depth level 3, which is classified as a truck, and resembles an animal at its upper half and a vehicle at its bottom, as can be seen in Fig. \ref{fig:f12-imaged_objects} (b). 
    \begin{figure}[htbp]
    \centering
    \includegraphics[width=0.75\textwidth]{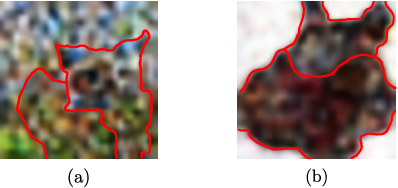}
    \caption{{\bf Examples of associated and imaged objects.} We draw a red contour to highlight important elements of the reconstructions.}
    \label{fig:f12-imaged_objects}
    \end{figure}
    \item Imaged objects assigned the correct class. For example: the frog, the truck and the airplane, indicating that such images are strong representative instances of their respective classes.
    \item Noisy images, which were often assigned the frog class. We reasoned that it is so because frogs usually have inherent noise formed by the variable patterns and pigmentation in their skin. 
\end{itemize}

The performance of the memory and the classifier in previous works were quite similar due in part to the simplicity of the images. However, the complexity of the images in the present experiment lowered the memory performance in relation to the classifier \cite{Hardman_2015}. Nevertheless, the accuracy of W-EAM on the \Cifar\ data set is comparable to the accuracy reported in \cite{Fagbohungbe_2022}. Furthermore, we chose from experiment 1 the AMR of size $1{,}024\times 16$ filled at 100\% of the \emph{Remembered Corpus} for the rest of the experiments. The compression ratio is 1,125 since the memory stores 12,000 images each of dimension 3,072, i.e., 36MB ---in a single register of 32KB.

The present investigation suggests that natural memory is accessed through cues very effectively, and that the distributed representation is highly efficient in relation to local schemes. Although retrieval is a reconstructive operation involving information loss, it has a satisfactory precision and recall trade-off.

\section*{Future work}
For future work we plan to examine the value of $\sigma$ and the response of the whole system with cues from highly unstructured corpora, such as: ImageNet \cite{deng_2009}, household audio \cite{belgoum_2018}, videos of human activity in the wild \cite{Soomro_2012} and social media applications in mental health\cite{Villa_2023}.

\section*{Acknowledgments}
The authors thank the members of the group {\em Memoria Asociativa y Racionalidad Entr\'opica} (MARE) for their constructive discussion and feedback.

\nolinenumbers


\begin{thebibliography}{10}

\bibitem{hopfield_1982}
Hopfield JJ.
\newblock Neural networks and physical systems with emergent collective
  computational abilities.
\newblock Proceedings of the National Academy of Sciences of the United States
  of America. 1982;79(8):2554--2558.

\bibitem{ritter_1999}
Ritter GX, de~Leon JLD, Sussner P.
\newblock Morphological bidirectional associative memories.
\newblock Neural Networks. 1999;12(6):851--867.

\bibitem{krotov_2016}
Krotov D, Hopfield JJ.
\newblock Dense Associative Memory for Pattern Recognition.
\newblock In: Lee D, Sugiyama M, Luxburg U, Guyon I, Garnett R, editors.
  Advances in Neural Information Processing Systems. vol.~29. Curran
  Associates, Inc.; 2016.

\bibitem{danihelka_2016}
Danihelka I, Wayne G, Uria B, Kalchbrenner N, Graves A.
\newblock Associative Long Short-Term Memory.
\newblock In: Proceedings of the 33rd International Conference on International
  Conference on Machine Learning - Volume 48. ICML'16. JMLR.org; 2016. p.
  1986–1994.

\bibitem{he_2019}
He H, Shang Y, Yang X, Di Y, Lin J, Zhu Y, et~al.
\newblock Constructing an Associative Memory System Using Spiking Neural
  Network.
\newblock Frontiers in Neuroscience. 2019;13.

\bibitem{le_2020}
Le H, Tran T, Venkatesh S.
\newblock Self-Attentive Associative Memory.
\newblock In: III HD, Singh A, editors. Proceedings of the 37th International
  Conference on Machine Learning. vol. 119 of Proceedings of Machine Learning
  Research. PMLR; 2020. p. 5682--5691.

\bibitem{pineda_2021}
Pineda LA, Fuentes G, Morales R.
\newblock An entropic associative memory.
\newblock Scientific Reports. 2021;11:6948.

\bibitem{morales_2022}
Morales R, Hernández N, Cruz R, Cruz VD, Pineda LA.
\newblock Entropic associative memory for manuscript symbols.
\newblock PLOS ONE. 2022;17(8):1--27.

\bibitem{lecun_2010}
LeCun Y, Cortes C, Burges C.
\newblock {MNIST} handwritten digit database.
\newblock ATT Labs [Online] Available: http://yannlecuncom/exdb/mnist. 2010;2.

\bibitem{cohen_2017}
Cohen G, Afshar S, Tapson J, van Schaik A. EMNIST: an extension of MNIST to
  handwritten letters; 2017.

\bibitem{pineda_img_2023}
Pineda LA, Morales R.
\newblock Imagery in the Entropic Associative Memory.
\newblock Scientific Reports. 2023;13:9553.

\bibitem{pineda_2022}
Pineda LA, Morales R.
\newblock Weighted entropic associative memory and phonetic learning.
\newblock Scientific Reports. 2022;12:16703.

\bibitem{pineda_2009}
Pineda LA, Castellanos H, Cu{\'{e}}tara J, Galescu L, Ju{\'{a}}rez J, Llisterri
  J, et~al.
\newblock The Corpus {DIMEx}100: transcription and evaluation.
\newblock Language Resources and Evaluation. 2009;44(4):347--370.

\bibitem{Zalando_2017}
Research Z. Fashion-MNIST; 2017.
\newblock Available from:
  \url{https://www.kaggle.com/datasets/zalando-research/fashionmnist}.

\bibitem{hinton_1986}
Hinton GE, Mcclelland JL, Rumelhart DE.
\newblock Distributed Representations.
\newblock In: Rumelhart DE, Mcclelland JL, editors. Parallel Distributed
  Processing: Explorations in the Microstructure of Cognition, {V}olume 1:
  {F}oundations. Cambridge, MA: MIT Press; 1986. p. 77--109.

\bibitem{Krizhevsky_2009}
Krizhevsky A.
\newblock Learning multiple layers of features from tiny images.
\newblock Canadian Institute for Advanced Research; 2009.

\bibitem{kozachkov_2023}
Kozachkov L, Slotine JJ, Krotov D.
\newblock Neuron-Astrocyte Associative Memory.
\newblock arXiv preprint arXiv:231108135. 2023;.

\bibitem{zhang_2023}
Zhang Y, Zeng Z.
\newblock Brain-Inspired Model and Neuromorphic Circuit Implementation for
  Feature-Affective Associative Memory Network.
\newblock IEEE Transactions on Cognitive and Developmental Systems. 2023; p.
  1--15.

\bibitem{jimnez_2023}
Jiménez M, Avedillo MJ, Linares-Barranco B, Núñez J.
\newblock Learning algorithms for oscillatory neural networks as associative
  memory for pattern recognition.
\newblock Frontiers in Neuroscience. 2023;17.

\bibitem{palm_1980}
Palm G.
\newblock On associative memory.
\newblock Biol Cybernetics. 1980;36:19--31.

\bibitem{willshaw_1969}
Willshaw D, Buneman O, Longuet-Higgins H.
\newblock Non-Holographic Associative Memory.
\newblock Nature. 1969;222:960--962.

\bibitem{simas_2023}
Simas R, Sa-Couto L, Wichert A. Classification and Generation of real-world
  data with an Associative Memory Model; 2023.

\bibitem{annabi_2022}
Annabi L, Pitti A, Quoy M.
\newblock On the relationship between variational inference and
  auto-associative memory.
\newblock In: Koyejo S, Mohamed S, Agarwal A, Belgrave D, Cho K, Oh A, editors.
  Advances in Neural Information Processing Systems. vol.~35. Curran
  Associates, Inc.; 2022. p. 37497--37509.

\bibitem{Raaijmakers_1980}
Raaijmakers JGW, Shiffrin RM.
\newblock SAM: A Theory of Probabilistic Search of Associative Memory.
\newblock vol.~14 of Psychology of Learning and Motivation. Academic Press;
  1980. p. 207--262.

\bibitem{MacKay_1991}
MacKay DJC.
\newblock In: Grandy WT, Schick LH, editors. Maximum Entropy Connections:
  Neural Networks. Dordrecht: Springer Netherlands; 1991. p. 237--244.

\bibitem{Sommer_1998}
Sommer FT, Dayan P.
\newblock Bayesian retrieval in associative memories with storage errors.
\newblock IEEE Transactions on Neural Networks. 1998;9(4):705--713.

\bibitem{Liu_2019}
Liu J, Gong M, He H.
\newblock Deep associative neural network for associative memory based on
  unsupervised representation learning.
\newblock Neural Networks. 2019;113:41--53.

\bibitem{Alvesdasilva_1992}
{Alves Da Silva} AP, Quintana VH, Pang GKH.
\newblock A probabilistic associative memory and its application to signal
  processing in electrical power systems.
\newblock Engineering Applications of Artificial Intelligence.
  1992;5(4):309--318.

\bibitem{janusz_2007}
Starzyk JA, He H, Li Y.
\newblock A Hierarchical Self-organizing Associative Memory for Machine
  Learning.
\newblock In: Liu D, Fei S, Hou ZG, Zhang H, Sun C, editors. Advances in Neural
  Networks -- ISNN 2007. Berlin, Heidelberg: Springer Berlin Heidelberg; 2007.
  p. 413--423.

\bibitem{Nakagawa_2006}
Nakagawa M.
\newblock Entropy Based Associative Model.
\newblock In: King I, Wang J, Chan LW, Wang D, editors. Neural Information
  Processing. Berlin, Heidelberg: Springer Berlin Heidelberg; 2006. p.
  397--406.

\bibitem{salvatori_2021}
Salvatori T, Song Y, Hong Y, Sha L, Frieder S, Xu Z, et~al.
\newblock Associative memories via predictive coding.
\newblock Neural Information Processing Systems Foundation; 2021.

\bibitem{kosko_2021}
Kosko B.
\newblock Bidirectional Associative Memories: Unsupervised Hebbian Learning to
  Bidirectional Backpropagation.
\newblock IEEE Transactions on Systems, Man, and Cybernetics: Systems.
  2021;51(1):103--115.
\newblock doi:{10.1109/TSMC.2020.3043249}.

\bibitem{hinton_2006}
Hinton GE, Salakhutdinov RR.
\newblock Reducing the Dimensionality of Data with Neural Networks.
\newblock Science. 2006;313(5786):504--507.

\bibitem{masci_2011}
Masci J, Meier U, Cire{\c{s}}an D, Schmidhuber J.
\newblock Stacked Convolutional Auto-Encoders for Hierarchical Feature
  Extraction.
\newblock In: Honkela T, Duch W, Girolami M, Kaski S, editors. Artificial
  Neural Networks and Machine Learning -- ICANN 2011. Berlin, Heidelberg:
  Springer Berlin Heidelberg; 2011. p. 52--59.

\bibitem{Fagbohungbe_2022}
Fagbohungbe O, Reza SR, Dong X, Qian L.
\newblock Efficient Privacy Preserving Edge Intelligent Computing Framework for
  Image Classification in IoT.
\newblock IEEE Transactions on Emerging Topics in Computational Intelligence.
  2022;6(4):941--956.

\bibitem{giuste_2020}
Giuste FO, Vizcarra JC. CIFAR-10 Image Classification Using Feature Ensembles;
  2020.

\bibitem{dosovitskiy_2021}
Dosovitskiy A, Beyer L, Kolesnikov A, Weissenborn D, Zhai X, Unterthiner T,
  et~al.
\newblock An Image is Worth 16x16 Words: Transformers for Image Recognition at
  Scale.
\newblock In: 9th International Conference on Learning Representations, {ICLR}
  2021, Virtual Event, Austria, May 3-7, 2021. OpenReview.net; 2021.

\bibitem{oquab_2023}
Oquab M, Darcet T, Moutakanni T, Vo H, Szafraniec M, Khalidov V, et~al..
  DINOv2: Learning Robust Visual Features without Supervision; 2023.

\bibitem{pineda_2024}
Pineda LA.
\newblock The mode of computing.
\newblock Cognitive Systems Research. 2024;84:101204.

\bibitem{Rishiraj_2013}
Mukherjee R, Islam T, Sankar R.
\newblock Text dependent speaker recognition using shifted MFCC.
\newblock In: 2013 Proceedings of IEEE Southeastcon; 2013. p. 1--4.

\bibitem{Padgett_1987}
Padgett RJ, Ratner HH.
\newblock Older and younger adults' memory for structured and unstructured
  events.
\newblock Experimental Aging Research. 1987;13(3):133--139.

\bibitem{Garland_1991}
Garland DJ, Barry JR.
\newblock Cognitive Advantage in Sport: The Nature of Perceptual Structures.
\newblock The American Journal of Psychology. 1991;104(2):211.
\newblock doi:{10.2307/1423155}.

\bibitem{Gobet_1996}
Gobet F, Simon HA.
\newblock Recall of random and distorted chess positions: Implications for the
  theory of expertise.
\newblock Memory {\&} Cognition. 1996;24(4):493--503.
\newblock doi:{10.3758/bf03200937}.

\bibitem{Buckley_2018}
Buckley BW, Daly L, Allen GN, Ridge CA.
\newblock Recall of structured radiology reports is significantly superior to
  that of unstructured reports.
\newblock The British Journal of Radiology. 2018; p. 20170670.
\newblock doi:{10.1259/bjr.20170670}.

\bibitem{Hardman_2015}
Hardman KO, Cowan N.
\newblock Remembering complex objects in visual working memory: Do capacity
  limits restrict objects or features?
\newblock Journal of Experimental Psychology: Learning, Memory, and Cognition.
  2015;41(2):325--347.

\bibitem{deng_2009}
Deng J, Dong W, Socher R, Li LJ, Li K, Fei-Fei L.
\newblock Imagenet: A large-scale hierarchical image database.
\newblock In: 2009 IEEE conference on computer vision and pattern recognition.
  IEEE; 2009. p. 248--255.

\bibitem{belgoum_2018}
Belgoum Z, Abdelnour J, Brodeur S, Rouat J. Audio dataset for Household
  Multimodal Environment (HoME); 2018.
\newblock https://dx.doi.org/10.21227/w77r-4902.

\bibitem{Soomro_2012}
Soomro K, Zamir AR, Shah M.
\newblock UCF101: A Dataset of 101 Human Actions Classes From Videos in The
  Wild.
\newblock CRCV-TR-12-01; 2012.

\bibitem{Villa_2023}
Villa-Pérez ME, Trejo LA. Twitter Dataset for Mental Disorders Detection;
  2023.
\newblock https://dx.doi.org/10.21227/6pxp-4t91.

\end{thebibliography}
%
%
%






\end{document}